\documentclass{article}

\usepackage{microtype}
\usepackage{graphicx}
\usepackage{subcaption}
\usepackage{booktabs}
\usepackage{url}
\usepackage{multirow}
\usepackage{enumitem}
\usepackage{graphicx}
\usepackage{balance}
\usepackage{booktabs} 
\usepackage{bm} 
\usepackage{hyperref}

\newcommand{\SC}{{\bf\texttt{ShapeCond}}}

\usepackage[preprint]{icml2026}
\usepackage{caption}
\usepackage{amsmath}
\usepackage{amssymb}
\usepackage{mathtools}
\usepackage{amsthm}

\usepackage[capitalize,noabbrev]{cleveref}

\theoremstyle{plain}

\theoremstyle{definition}

\theoremstyle{remark}

\usepackage[textsize=tiny]{todonotes}

\icmltitlerunning{ShapeCond: Fast Shapelet-Guided Dataset Condensation}

\begin{document}

\twocolumn[
  \icmltitle{\SC: Fast {\em Shapelet}-Guided Dataset Condensation \\ for Time Series Classification}

  \icmlsetsymbol{equal}{*}

  \begin{icmlauthorlist}
    \icmlauthor{Sijia Peng}{fudan,MBZU}
    \icmlauthor{Yun Xiong}{fudan}
    \icmlauthor{Xi Chen}{fudan}
    \icmlauthor{Yi Xie}{fudan}
    \icmlauthor{Guanzhi Li}{stanford}
    \icmlauthor{Yanwei Yu}{oc}
    \icmlauthor{Yangyong Zhu}{fudan,DRI}
    \icmlauthor{Zhiqiang Shen}{MBZU}
  \end{icmlauthorlist}

  \icmlaffiliation{fudan}{Shanghai Key Lab of Data Science, College of Computer Science and Artificial Intelligence, Fudan University, Shanghai, China}
  \icmlaffiliation{stanford}{Stanford University, Stanford, USA}
  \icmlaffiliation{oc}{Ocean University of China}
  \icmlaffiliation{DRI}{Shanghai Data Research Institute}
  \icmlaffiliation{MBZU}{Mohamed bin Zayed University of Artificial Intelligence
Abu Dhabi, United Arab Emirates}

  \icmlcorrespondingauthor{Yun Xiong}{yunx@fudan.edu.cn}
  \icmlcorrespondingauthor{Zhiqiang Shen}{zhiqiang.shen@mbzuai.ac.ae}

  \icmlkeywords{Machine Learning, Dataset Condensation}

  \vskip 0.3in
]

\printAffiliationsAndNotice{}

\begin{abstract}
Time series data supports many domains (e.g., finance and climate science), but its rapid growth strains storage and computation. Dataset condensation can alleviate this by synthesizing a compact training set that preserves key information. Yet most condensation methods are image-centric and often fail on time series because they miss time-series-specific temporal structure, especially local discriminative motifs such as shapelets.
In this work, we propose \SC{}, a novel and efficient condensation framework for time series classification that leverages {\em shapelet}-based dataset knowledge via a {\em shapelet}-guided optimization strategy. Our shapelet-assisted synthesis cost is independent of sequence length: longer series yield larger speedups in synthesis (e.g., {\bf 29$\times$} faster over prior state-of-the-art method CondTSC for time-series condensation, and up to \textbf{10,000$\times$} over naively using shapelets on the Sleep dataset with 3,000 timesteps). By explicitly preserving critical local patterns, \SC{} improves downstream accuracy and consistently outperforms all prior state-of-the-art time series dataset condensation methods across extensive experiments.
Code is available at \url{https://github.com/lunaaa95/ShapeCond}.
\end{abstract}

\section{Introduction}
\label{sec:intro}

\begin{figure}[t]
\centering
\includegraphics[width=0.9\linewidth]{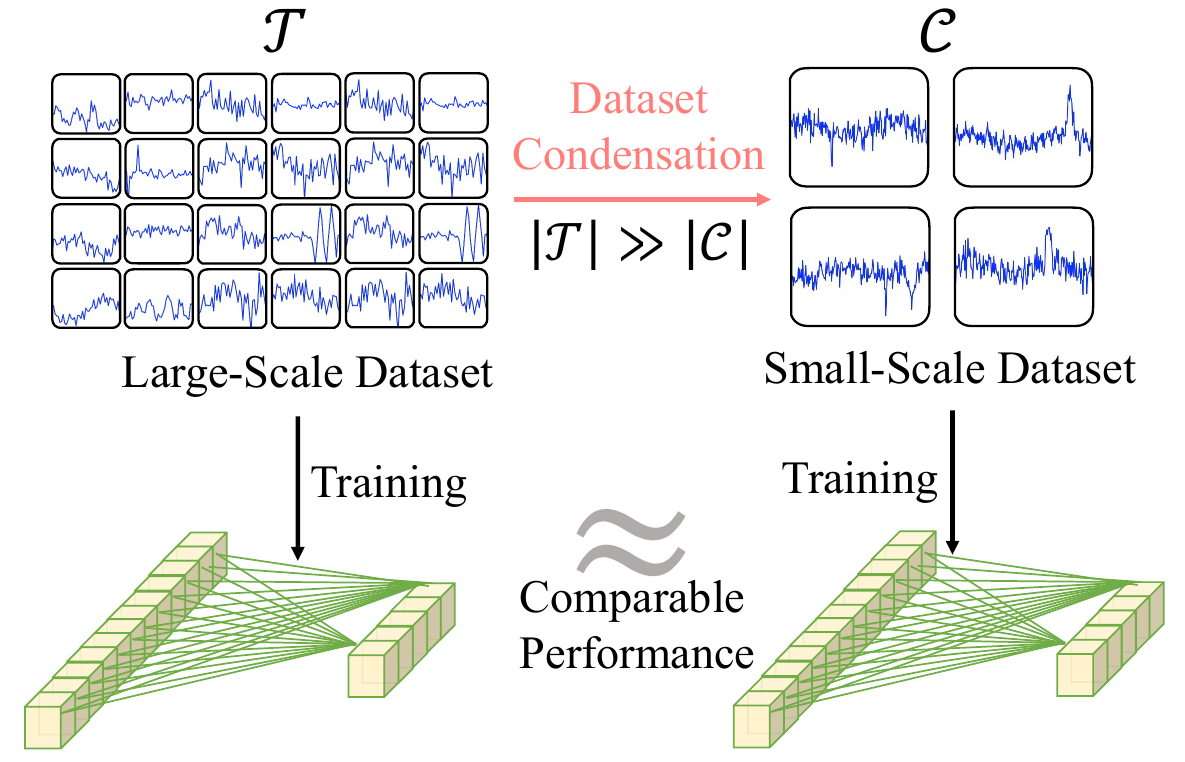}
\caption{Conceptual illustration of dataset condensation. The aim is to synthesize a small yet informative dataset from a large dataset, such that models trained on the condensed dataset can achieve performance comparable to those trained on full dataset. This approach facilitates faster training and reduces storage requirements.}
\label{fig: task}
\vspace{-0.1in}
\end{figure}

Time series data has been extensively investigated for various applications, such as fraud detection \cite{rousseeuw2019robust, devaki2014credit, kemp2021empty}, human activity recognition \cite{yang2015deep, alawneh2021enhancing}, and disease diagnosis \cite{song2018attend, bui2018time}, where models must learn from long, high-frequency sequences. As data volumes grow, training and iterating on full datasets becomes increasingly expensive, both in storage and in compute, slowing experimentation and deployment. Dataset condensation \cite{wang2018dataset, zhao2023dataset, sre2l} offers an appealing alternative: synthesize a small set of representative training sequences that can replace the original data while preserving downstream accuracy as shown in Fig.~\ref{fig: task}. However, most condensation methods were developed for images and do not transfer well to time series, where discriminative information is often carried by temporal dynamics and localized segments rather than spatial textures.

\begin{figure}[!t]
\centering
\includegraphics[width=0.99\linewidth]{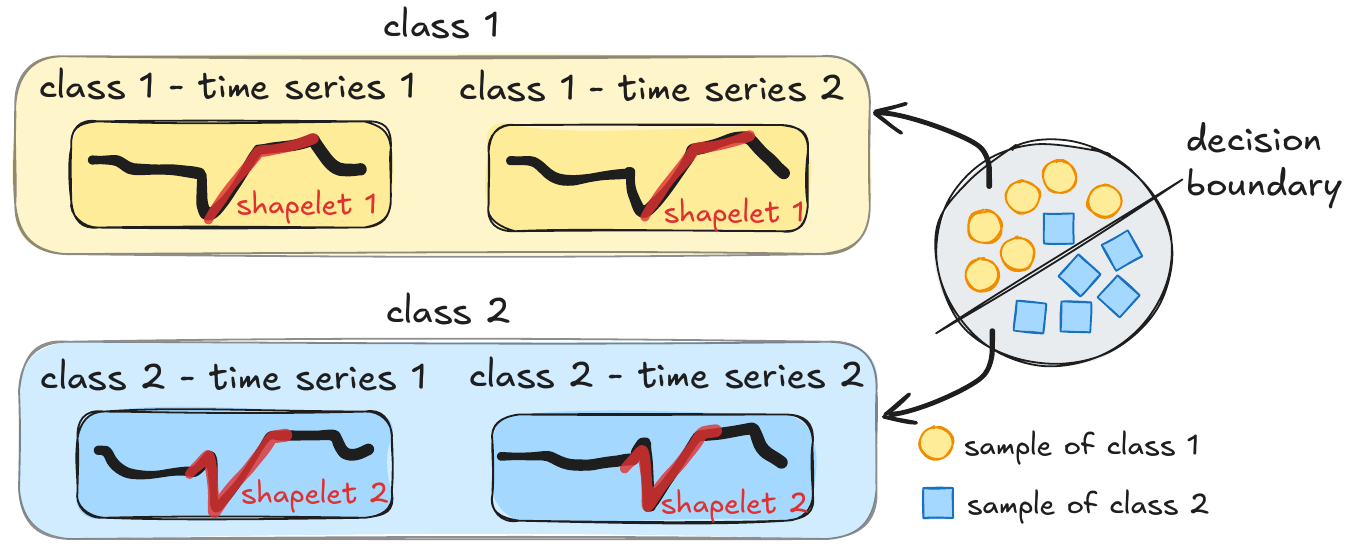}
\caption{Shapelets are the most representative segments within classes. Thus, the classification of a time series with an unknown label can be determined by comparing its Euclidean distance to shapelet 1 (for class 1) with that to shapelet 2 (for class 2).}
\label{fig: shapelets}
\end{figure}

A central challenge is that time series contain structure at multiple scales. Global trends and long-range dependencies capture coarse class signatures (e.g., circadian rhythm patterns), while local discriminative subsequences, classically studied as {\em shapelets}, often determine the decision boundary (e.g., a brief arrhythmic segment in electrocardiogram) as shown in Fig.~\ref{fig: shapelets}. Existing condensation pipelines tend to emphasize global matching objectives (e.g., feature statistics~\cite{sre2l} or training trajectories~\cite{mtt}) and can miss these localized segments, leading to condensed sets that are compact yet semantically impoverished for time-series decision rules. Moreover, many approaches~\cite{condtsc,condtsf} incur costs that scale with sequence length, making condensation particularly slow for long sequences.

We introduce {\SC}, an efficient {\em shapelet}-guided dataset condensation framework for time series classification. ShapeCond is built on the observation that shapelets encode intrinsic dataset knowledge: they serve as compact, class-representative ``evidence snippets'' that a classifier can use to discriminate categories via similarity in the temporal domain. Fig.~\ref{fig: shapelets} illustrates the discriminative power of shapelets. Instead of treating time series as generic vectors, ShapeCond explicitly incorporates shapelets as first-class guidance for synthesis, enabling the condensed set to retain the most salient local patterns while remaining lightweight. This design directly addresses the key failure mode of image-centric condensation methods on temporal data.

At the core of ShapeCond is a novel {\em Global–Local Temporal Structure Optimization} mechanism that jointly preserves (i) global temporal structure to capture long-range dynamics and overall sequence morphology, and (ii) local shapelet structure to ensure the condensed sequences contain the discriminative motifs that drive classification. Concretely, ShapeCond first identifies a set of informative shapelets from the full dataset and then optimizes condensed sequences that are close to these shapelets to remain faithful to these motifs while maintaining coherent global dynamics. By coupling global alignment with explicit local motif preservation, ShapeCond avoids collapsing to overly smooth prototypes and prevents the loss of short yet decisive temporal evidence.
Beyond accuracy, ShapeCond is designed for practical efficiency at long horizons. Our synthesis procedure has computational cost that is effectively independent of sequence length, so increasing timesteps does not slow down the core optimization. Longer series amplify the advantage over prior baselines whose costs scale linearly (or worse) with the number of timesteps. 

Extensive experiments demonstrate that ShapeCond consistently delivers state-of-the-art performance across standard time series classification benchmarks while substantially reducing storage and training cost. We show that models trained on the condensed set match those trained on competing condensed datasets, particularly in settings where localized motifs are essential. Ablations further validate the complementary roles of global and local objectives. Moreover, ShapeCond achieves up to a {\bf 29$\times$ speedup}\footnote{Up to 10,000× over naively using shapelets on the Sleep dataset with 3,000 timesteps. Detailed analysis is in Appendix~\ref{app_cost}.} with state-of-the-art accuracy, enabling condensation to be used routinely even for high-resolution physiological recordings.

Our contributions are summarized as follows:

\begin{itemize}
    \item We propose an efficient {\em shapelet}-guided optimization approach that integrates shapelets into the dataset condensation process. We highlight the intrinsic dataset knowledge inherent in shapelets, and leverage these local discriminative patterns, which have been proven critical for fast and accurate modeling.
    
    \item We introduce \SC, a novel dataset condensation framework for time series data. Our method adopts a dual-view data synthesis paradigm, where global temporal structure is controlled by model encoder gradients, while the local key patterns are refined by {\em shapelet}-guided optimization. This enables a comprehensive preservation of original temporal dynamics, enhancing condensation performance for time series. 

    \item We conduct extensive experiments on seven public datasets at both small and large scales to evaluate the effectiveness of our proposed method. Our method outperforms strong prior baselines by 17.56\% on average, achieving state-of-the-art results across all datasets.
\end{itemize}

\begin{figure*}
\centering
\includegraphics[width=0.9\linewidth]{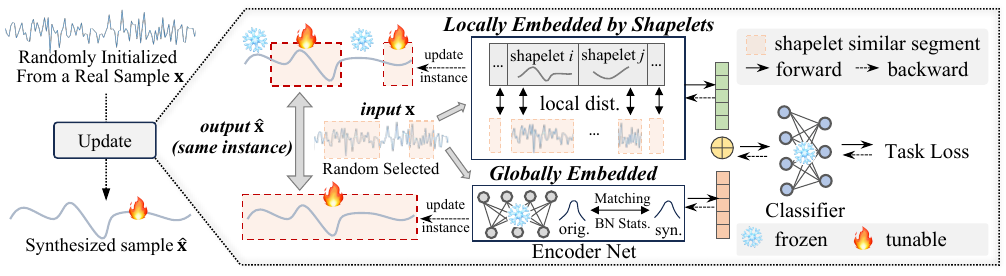}
\caption{Shapelet-guided Data Synthesis Stage. While model encoder gradients control global temporal structure (lower box), {\em shapelet}-guided optimization preserves critical local patterns (upper box).}
\label{fig: optmization}
\vspace{-0.1in}
\end{figure*}

\vspace{-0.1in}
\section{Problem Statement}

We first present two concepts central to this work: (i) the task of \emph{dataset condensation} on time series dataset format, and (ii) the data modality of \emph{time series}.

\subsubsection*{Task: Dataset Condensation} 
Given a dataset $\mathcal{T} = (X,Y) = \{{\bf x}_{i}, y_{i}\}_{i=1}^{|\mathcal{T}|}$, where each $({\bf x}_i, y_i)$ is a sample pair in $\mathcal{T}$, the goal of dataset condensation is to synthesize a significantly smaller dataset $\mathcal{C} = (\hat{X}, \hat{Y}) = \{\hat{\bf x}_{i}, \hat{y}_{i}\}_{i=1}^{|\mathcal{C}|}$ that retains the essential information of $\mathcal{T}$. The condensed dataset $\mathcal{C}$ is expected to satisfy the following properties:

\begin{enumerate}
    \item The size of $\mathcal{C}$ is substantially smaller than that of $\mathcal{T}$, i.e., $|\mathcal{C}| \ll |\mathcal{T}|$.
    \item Models trained on $\mathcal{C}$ should achieve comparable performance to those trained on $\mathcal{T}$. Formally, let
    \begin{equation}
        \theta_{\mathcal{C}} = \underset{\theta}{\arg \min} \sum_{(\hat{\bf x}_i, \hat{y}_i) \in \mathcal{C}} \ell\left(\hat{\bf x}_i, \hat{y}_i, \mathcal{F}_\theta\right),
    \end{equation}
    \begin{equation}
        \theta_{\mathcal{T}} = \underset{\theta}{\arg \min} \sum_{({\bf x}_i, y_i) \in \mathcal{T}} \ell\left({\bf x}_i, y_i, \mathcal{F}_\theta\right),
    \end{equation}
\end{enumerate}
 Then we expect their performance to satisfy:
    \begin{equation}
    \hspace{-3em}
    \begin{aligned}
        \mathbb{E}_{({\bf x}, y) \in \mathcal{T}_{\text{test}}} \left[\operatorname{eval}\left(\mathcal{F}_{\theta_{\mathcal{C}}}({\bf x}), y\right)\right] 
        \simeq \\ 
        \mathbb{E}_{({\bf x}, y) \in \mathcal{T}_{\text{test}}} \left[\operatorname{eval}\left(\mathcal{F}_{\theta_{\mathcal{T}}}({\bf x}), y\right)\right],
    \end{aligned}
    \end{equation}
where $\ell\left(\cdot, \cdot\right)$ is training loss, $\mathbb{E}\left[\cdot\right]$ denotes the expectation, 
$\mathcal{F}_\theta$ is model $\mathcal{F}$ parameterized by $\theta$, and 
$\operatorname{eval}$ is the metric to evaluate the model performance on the test set, and

The dataset $\mathcal{C}$ is synthesized from $\mathcal{T}$ by an algorithm $\mathcal{A}$. In this work, $\mathcal{A}$ refers to our proposed method, \SC, which is specifically designed for time series dataset condensation. By generating $\mathcal{C}$, we aim to significantly reduce storage requirements and training costs while preserving model performance.

\subsubsection*{Data Type: Time Series}

We focus on time series classification datasets, where each example is a pair $({\bf x}_i,y_i)$ consisting of a time series ${\bf x}_i$ and a class label $y_i$.
A (univariate) time series is an ordered sequence of real-valued observations indexed by time, typically represented as ${\bf x}\in\mathbb{R}^{L}$, where $L$ is sequence length.

Unlike images or text, time series present unique properties for condensation. Discriminative information is often concentrated in short, localized subsequences (motifs), where temporal positions can be important. Meanwhile, long-range temporal dependencies, such as trends or cycles, must also be preserved. These characteristics make naive condensation approaches challenging, motivating a framework that maintains both local motifs and global temporal structure.

\vspace{-0.1in}
\section{Approach}

In this section, we introduce \SC{} in a step-by-step manner, an efficient framework designed for time-series dataset condensation. \SC{} explicitly disentangles local and global temporal information and jointly optimizes these two complementary components. The key idea is to synthesize condensed data by optimizing two views in parallel: (i) global temporal structure and (ii) localized, task-critical patterns, and then integrate them within a unified optimization framework.

To enable this design under the large-scale setting of dataset condensation, we reformulate classical shapelet discovery into a scalable procedure, making shapelet-based modeling of informative local motifs computationally feasible.

The condensed dataset is synthesized via a dual-view optimization process, where global temporal dynamics and local pattern constraints are jointly enforced. This complementary optimization allows the condensed data to preserve both long-range dynamics and critical local structures, leading to more faithful dataset compression.
\SC{} proceeds in three stages: \textit{Fast Shapelet Discovery}, \textit{Knowledge Fetching}, and \textit{Data Synthesis}.

\subsection{Fast \emph{Shapelet} Discovery}
\label{sec:fast_shapelet}

Since our goal is to make the proposed method scalable, we redesign classical shapelet discovery into a structure-aware procedure. Instead of extracting shapelets primarily or solely for instance-level classification, we identify dataset-level, discriminative local patterns that can efficiently guide synthetic data generation. This subsection details the objective and motivates the proposed pipeline.

\begin{figure}[t]
\centering
\includegraphics[width=0.99\linewidth]{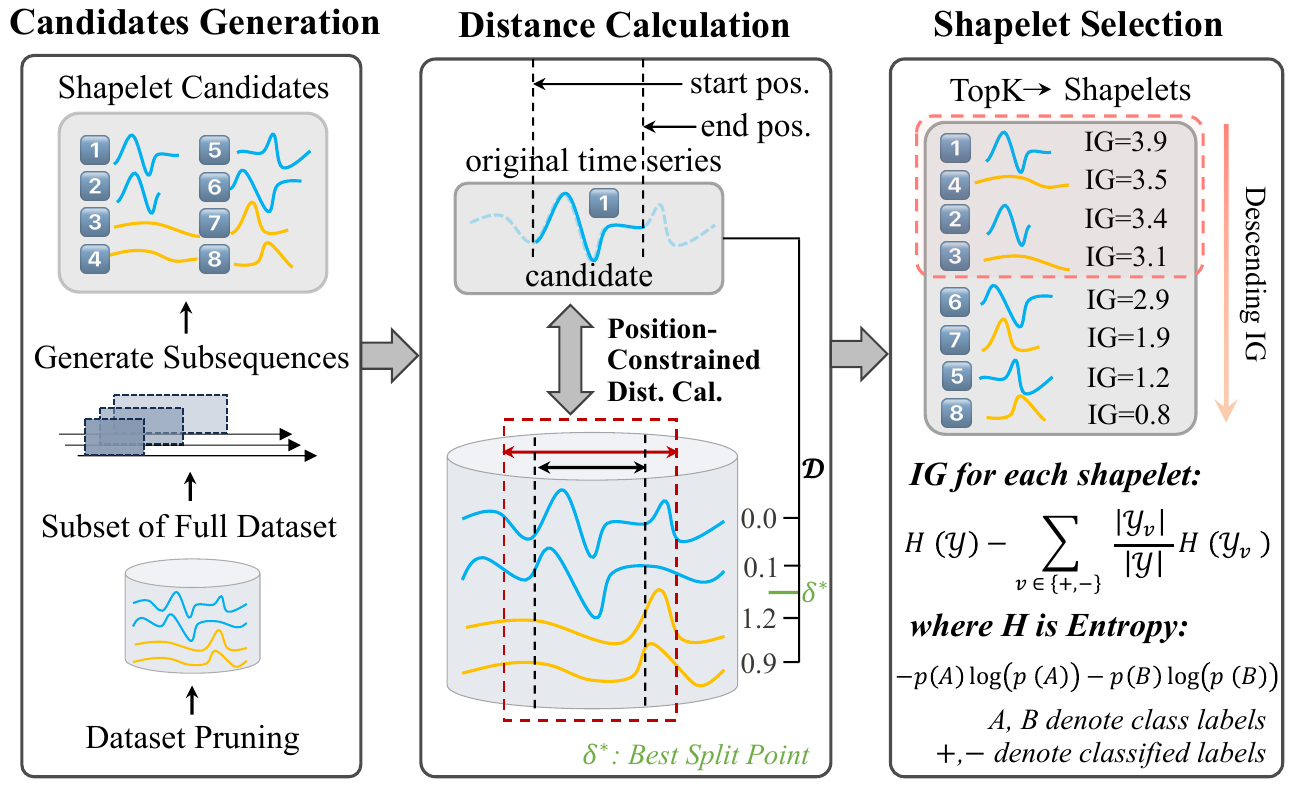}
\caption{Fast Shapelet Discovery.
We reduce the computational cost by combining candidate pruning with position-constrained distance search. 
Distance computation is restricted to a local temporal neighborhood, including the candidate’s original position (red double arrows) and nearby temporal positions (black double arrows), yielding constant-time ($O(1)$) distance evaluations.}
\label{fig: shapelet generation}

\end{figure}

\subsubsection{Our Pipeline}
Guided by the above design motivation, our fast shapelet discovery pipeline consists of three steps:

\textbf{Candidate Generation with Data Pruning.}
A candidate shapelet is defined as a contiguous subsequence:
\vspace{-0.05in}
\begin{equation}
\mathbf{s}_{i,j,l} \!=\! (x_{ij}, x_{i(j+1)}, \dots, x_{i(j+l-1)}), 
\quad 0 \le j \le L-l.
\end{equation}
Let $|\mathcal{T}| = N$ and let $M$ denote the number of candidate shapelets
under standard enumeration.
We generate candidates from a randomly pruned subset
$\mathcal{T}' \subset \mathcal{T}$ with $(1-p)N$ samples,
where $p \in (0,1)$ is the pruning ratio,
resulting in $(1-p)M$ candidates.

\textbf{Position-Constrained Distance Evaluation.}
For a candidate shapelet $\mathbf{s}$ extracted at temporal position $j$, the distance to a time series ${\bf x}_i$ is classically defined as:
\begin{equation}
D(\mathbf{s}, {\bf x}_i)
= \min_{\mathbf{s}' \in \{\mathbf{s}_{i,j',|\mathbf{s}|}\}}
Dist(\mathbf{s}, \mathbf{s}'),
\end{equation}
where $Dist(\cdot)$ denotes the Euclidean distance.
To achieve scalable computation, we restrict the search to a fixed-size local temporal neighborhood $\mathcal{N}(j)$ around the original position:
\begin{equation}
\tilde{D}(\mathbf{s}, {\bf x}_i)
= \min_{j' \in \mathcal{N}(j)}
Dist(\mathbf{s}, \mathbf{s}_{i,j',|\mathbf{s}|}).
\end{equation}
This position-constrained evaluation removes the dependence on the sequence length and enables constant-time distance computation.

\textbf{Shapelet Scoring and Selection.}
Each candidate shapelet $\mathbf{s}$ is evaluated using information gain.  
Given a threshold $\delta$, instances are partitioned as:
\[
\mathcal{Y}_v = \{ y_i \mid \tilde{D}(\mathbf{s}, {\bf x}_i) \le \delta \}, \quad v \in \{+,-\},
\]
and the corresponding information gain is:
\begin{equation}
IG(\mathbf{s}, \delta)
= H(\mathcal{Y}) -
\sum_{v \in \{+,-\}}
\frac{|\mathcal{Y}_v|}{|\mathcal{Y}|} H(\mathcal{Y}_v),
\end{equation}
where the entropy is defined as
$H(\mathcal{Y}) = -p(A)\log p(A) - p(B)\log p(B)$.
The discriminative power of $\mathbf{s}$ is given by its best split:
\begin{equation}
IG(\mathbf{s}) = \max_{\delta} IG(\mathbf{s}, \delta).
\end{equation}
The top-$k$ candidates with the highest information gain are selected to form the shapelet pool
$\mathcal{S}^* = \{\mathbf{s}_1, \dots, \mathbf{s}_k\}$.
Any time series ${\bf x}$ is then transformed into a shapelet-based representation:
\begin{equation}
\label{eq:sl_trans}
STrans({\bf x};\mathcal{S}^*)
= \big[ \tilde D({\bf x}, \mathbf{s}_1), \dots, \tilde D({\bf x}, \mathbf{s}_k) \big],
\quad \mathbf{s}_i \in \mathcal{S}^*.
\end{equation}

\subsubsection{Computational Complexity Analysis}
The efficiency of our fast shapelet discovery is achieved through two key strategies:

\textbf{(1) Efficient Subsampling via Redundancy.} 
Time series datasets often contain redundant samples. 
We perform shapelet discovery on a randomly selected subset of training data. 
As shown in Sec.~\ref{sec:PR}, pruning 50\% of samples ($p=0.5$) causes no accuracy drop, and pruning up to 70\% ($p=0.7$) preserves accuracy in most cases, 
while reducing computational cost to roughly $(1-p)^2$ of the original.

\textbf{(2) Position-Constrained Distance Search.} 
Instead of computing the minimum distance between each candidate and all subsequences, 
we restrict the search to a local temporal neighborhood around the candidate’s original position as shown in Fig.~\ref{fig: shapelet generation}. 
This reduces distance evaluations substantially while preserving shapelet quality, making the computation independent of the sequence length $L$. 

\begin{table}[h!]
\centering
\caption{Comparison of computational complexity between standard shapelet discovery~\cite{shapelets} and our proposed method in \SC{}. 
Dis.\ cal. refers to distance calculation.}
\label{tab:complexity_comparison}
\resizebox{0.48\textwidth}{!}{
\begin{tabular}{cccc}
\toprule
Method & \# Candidates & Cost per Dis.\ cal. & Dis.\ cal. Times \\
\midrule
Standard & $M$ & $O(L)$ & $M \cdot N$ \\
Our Proposed & $M \cdot (1-p)$ & $O(1)$ & $(1-p)^2 \cdot M \cdot N$ \\
\bottomrule
\end{tabular}
}
\end{table}

Here, $N$ is the number of time series, $L$ is the sequence length, $M$ is the number of candidate shapelets retained after pruning, and $L$ is the sequence length.
Notably, our distance computation is \emph{independent of sequence length} $L$, yielding substantial savings, especially for longer sequences. 

Table~\ref{tab:complexity_comparison} summarizes the reduction in computational cost achieved by our method, including the number of candidates, the cost per distance evaluation, and the number of distance evaluations required. A detailed theoretical efficiency analysis is provided in Appendix~\ref{app:complexity_shapelet}.

\subsection{Knowledge Fetching}

This stage constructs a teacher model that captures dataset knowledge and stores it into model parameters. 

We maintain the \textit{framework} and \textit{size of parameters} of teacher model, but slightly modify the forward process to allow shapelet-transformed input.

Specifically, given a model $\mathcal{F}_{\theta_{\mathcal{T}}}$ composed of an encoder $f_\text{enc}$ and a classifier $f_\text{cls}$, parameterized by ${\theta_{\mathcal{T}}}$, trained on full dataset $\mathcal{T}$, and a set of representative shapelets ${\bf \mathcal{S}}^*$, we concatenate the representations from encoder and shapelet transformation to generate final classification. This forward process of teacher model is expressed by:
\begin{equation}
    {\bf y_{\text{pred}}}  = \mathcal{F}_{\theta_{\mathcal{T}}}({\bf x};\mathcal{S}^*) = f_\text{cls}([f_\text{enc}({\bf x}), STrans({\bf x};\mathcal{S}^*)]),
\end{equation}
${\bf y_{\text{pred}}} \in \mathbb{R}^V$ is the predicted class distribution with $V$ being the number of classes, $f_\text{enc}({\bf x}$) contains knowledge learned by encoder net, $STrans({\bf x};\mathcal{S}^* )$ carries shapelet knowledge.

Then, following standard training process, teacher model is trained on $\mathcal{T}$ to obtain optimal model parameters $\theta_{\mathcal{T}}$:
\begin{equation}
    \theta_{\mathcal{T}} = \underset{\theta}{\arg \min} \sum_{({\bf x}, y) \in \mathcal{T}} \ell\left(\mathcal{F}_{\theta_{\mathcal{T}}}({\bf x};\mathcal{S}^*), y\right),
\end{equation}
where $\ell$ is loss function, typically cross-entropy loss:
\begin{equation}
\ell\left(\mathcal{F}_{\theta_{\mathcal{T}}}({\bf x};\mathcal{S}^*), y\right) = -\log \left( {\bf \mathcal{F}_{\theta_{\mathcal{T}}}({\bf x};\mathcal{S}^*)}^{(y)} \right).
\end{equation}
where ${\bf \mathcal{F}_{\theta_{\mathcal{T}}}({\bf x};\mathcal{S}^*)}^{(y)}$ is the predicted probability for the true class $y$, i.e., the $y_i$-th element in the predicted vector ${\bf \mathcal{F}_{\theta_{\mathcal{T}}}({\bf x};\mathcal{S}^*)}^{(y)}$.

In the following stage, this trained teacher model ${\theta}_\mathcal{T}$ is expected to guide the synthesis of condensed dataset.

\subsection{Data Synthesis via Dual Optimization}

At this stage, we synthesize the condensed dataset by initializing a small labeled set and optimizing it into an informative proxy of the full data as shown in Fig.~\ref{fig: optmization}.

\noindent\textbf{Initialization.}  We initialize the condensed data $\mathcal{C}$ from randomly selected real samples in $\mathcal{T}$. The condensed data covers all classes in the original dataset with an equal number of samples per class to ensure class balance. 

\noindent\textbf{Global--Local Temporal Structure Optimization.}
We adopt a model inversion paradigm: a well-trained teacher $\mathcal{F}_{\theta_{\mathcal{T}}}$ is frozen and used to guide the optimization of $\mathcal{C}$ through task supervision and distribution statistics. The key novelty of \SC{} is to integrate \emph{shapelet-guided} optimization into condensation. Concretely, global temporal structure is learned via the teacher encoder, while critical local patterns are enforced through a shapelet transform module, encouraging the condensed set to preserve essential shapelet signatures (see Fig.~\ref{fig: optmization}). Our objective is:
\begin{equation}
\mathcal{C}=\arg\min_{\mathcal{C}} \sum_{(\hat{\bf x},\hat{y})\in\mathcal{C}}
\ell\!\left(\mathcal{F}_{\theta_{\mathcal{T}}}((\hat{\bf x};\mathcal{S}^*)),\hat{y}\right)
+\mathcal{R}_{db}(\mathcal{C}),
\end{equation}
where the task loss is back-propagated through both the encoder and the shapelet transform, yielding a dual-view update:
(i) global dynamics from the encoder, and (ii) local discriminative motifs from the shapelet transform. This ensures $\mathcal{C}$ retains both comprehensive temporal structure and localized shapelet patterns.

\noindent\textbf{Statistics matching.}
We also align the distributions of $\mathcal{T}$ and $\mathcal{C}$ by matching BatchNorm statistics:
\begin{equation}
\mathcal{R}_{db}(\mathcal{C})=\sum_l \Big(
\|\mu_l(\hat{X})-\mu_l^{BN}\|_2^2+
\|\sigma_l^2(\hat{X})-\sigma_l^{BN}\|_2^2
\Big).
\end{equation}
where $\hat{X}$ is the set of all condensed instances, and $\mu_l(\hat{X}),\sigma_l^2(\hat{X})$ are the mean/variance at the $l$-th BN layer, with $\mu_l^{BN},\sigma_l^{BN}$ the teacher's running statistics.

After optimization, we replace hard labels with teacher soft labels to increase informativeness, yielding
$\mathcal{C}=\{(\hat{\bf x}_i,\hat{\bf y}^{\text{soft}}_i)\}$ with
$\hat{\bf y}^{\text{soft}}_i=\mathcal{F}_{\theta_{\mathcal{T}}}((\hat{\bf x}_i;\mathcal{S}^*))$.

\begin{table*}[tp]
\centering
\caption{
Overall performance (Accuracy \%). ``Ratio'' refers to the condensed ratio, calculated as $\frac{\text{size of condensed dataset}}{\text{size of full dataset}}$, and ``SPC'' denotes the number of samples per class in the condensed dataset. ``Full'' represents the performance of a model trained on the full dataset, serving as the upper bound. All the reported values are averaged over three experimental runs. Best accuracy highlighted in bold.}

\label{t:main results}
\resizebox{.93\textwidth}{!}{
\begin{tabular}{lcccccccccccc}
\toprule
Dataset & Ratio(\%) & SPC & Random & Herding & K-Center & DC & DSA & MTT & SRe$^2$L & CondTSC & \SC{} & Full \\
\midrule
\multirow{3}{*}{FacesUCR} & 1 & 1 & 31.63 & 39.87 & 40.76 & 58.23 & 56.46 & 68.47 & 59.30 & 67.75 & \textbf{79.26} & \multirow{3}{*}{97.12} \\
 & 5 & 5 & 70.82 & 64.14 & 68.82 & 75.81 & 76.98 & 86.96 & 92.78 & 88.69 & \textbf{93.26} &  \\
 & 10 & 10 & 77.73 & 77.95 & 77.51 & 80.06 & 84.66 & 92.42 & 94.09 & 92.65 & \textbf{95.10} &  \\
\midrule
\multirow{2}{*}{TP} & 0.3 & 1 & 24.91 & 24.00 & 24.00 & 33.49 & 33.14 & 42.79 & 33.42 & 43.79 & \textbf{45.95} & \multirow{2}{*}{99.95} \\
 & 3.2 & 10 & 39.84 & 34.82 & 43.25 & 49.18 & 50.24 & 75.35 & 77.08 & 78.40 & \textbf{90.94} &  \\
\midrule
\multirow{3}{*}{HAR} & 0.1 & 1 & 52.58 & 40.59 & 38.11 & 52.50 & 53.80 & 60.54 & 60.12 & 62.45 & \textbf{76.44} & \multirow{3}{*}{95.73} \\
 & 0.5 & 5 & 54.62 & 62.43 & 64.10 & 64.90 & 66.47 & 81.63 & 79.69 & 82.20 & \textbf{89.10} &  \\
 & 1.0 & 10 & 62.27 & 65.32 & 66.51 & 69.65 & 72.07 & 90.06 & 84.83 & 82.68 & \textbf{92.08} &  \\
\midrule
\multirow{3}{*}{Electric} & 0.1 & 1 & 39.29 & 39.55 & 42.86 & 45.12 & 46.06 & 52.06 & 49.95 & 45.32 & \textbf{54.72} & \multirow{3}{*}{75.20} \\
 & 0.4 & 5 & 38.44 & 48.93 & 46.79 & 53.74 & 56.91 & 60.03 & 62.53 & 54.24 & \textbf{65.38} &  \\
 & 0.9 & 10 & 46.48 & 58.96 & 47.78 & 55.51 & 55.34 & 62.89 & 63.87 & 56.52 & \textbf{68.30} &  \\
\midrule
\multirow{2}{*}{Sleep} & 0.2 & 10 & 44.54 & 42.30 & 32.09 & 32.16 & 32.10 & 35.77 & 29.43 & 46.16 & \textbf{47.08} & \multirow{2}{*}{75.53} \\
 & 1.0 & 50 & 52.62 & 45.23 & 50.31 & 42.35 & 42.67 & 54.28 & 63.86 & 60.27 & \textbf{68.21} &  \\
\midrule
\multirow{3}{*}{Tiselac} & 0.11 & 10 & 59.85 & 63.24 & 64.03 & 61.58 & 63.41 & 62.88 & 47.42 & 61.49 & \textbf{72.83} & \multirow{3}{*}{80.60} \\
 & 0.23 & 20 & 65.17 & 67.00 & 63.26 & 62.28 & 68.82 & 68.86 & 50.55 & 63.18 & \textbf{75.37} &  \\
 & 0.56 & 50 & 71.97 & 73.71 & 72.54 & 69.71 & 70.02 & 72.64 & 61.04 & 70.52 & \textbf{77.18} &  \\
 \midrule
\multirow{4}{*}{Pedestrian} & 0.05 & 1 & 6.98 & 4.29 & 4.95 & 3.77 & 6.77 & 6.40 & 8.92 & 10.29 & \textbf{10.78} & \multirow{4}{*}{31.37} \\
 & 0.27 & 5 & 11.56 & 10.55 & 10.24 & 5.53 & 9.29 & 12.68 & 17.50 & 17.20 & \textbf{25.30} &  \\
 & 0.54 & 10 & 11.85 & 15.99 & 13.52 & 7.18 & 12.56 & 15.09 & 20.52 & 17.89 & \textbf{27.69} &  \\
 & 1.08 & 20 & 13.10 & 16.22 & 21.49 & 9.81 & 16.44 & 18.43 & 27.41 & 21.00 & \textbf{30.00} & \\
\bottomrule
\end{tabular}
}
\vspace{-0.1in}
\end{table*}

\vspace{-0.05in}
\section{Experiments}
In this section, we present extensive experiments to evaluate the effectiveness of our proposed \SC~for time series dataset condensation.

\subsection{Setups}
\noindent{\bf Datasets.} Experiments are conducted on seven public time-series datasets: FacesUCR, TwoPatterns, HAR, ElectricDevices, Sleep, Tiselac, and Pedestrian. All models are implemented in PyTorch and conducted on four NVIDIA V100 32GB GPUs.

\noindent{\bf Optimizer.} We use AdamW optimizer with $1e^{-4}$ weight decay in training of Knowledge Fetching stage and the evaluation of synthesized dataset (Evaluation). 
The default learning rate is set to $1e^{-4}$ (searching space ranges from $1e^{-3}$ to $1e^{-6}$) for Knowledge Fetching stage, and $1e^{-3}$ for Evaluation (searching space ranges from $1e^{-1}$ to $1e^{-4}$). 
For Data Synthesis stage, we use Adam optimizer with $betas=[0.5, 0.9]$, and learning rate is set to $0.2$ by default (searching space ranges from $0.1$ to $0.4$).

\noindent{\bf Backbone.}\label{sec:backbone} Following previous dataset condensation works that use simple architectures as backbones to evaluate the performance of condensation methods, our evaluation is conducted on widely used convolutional neural networks (CNNs). Specifically, except for Section~\ref{sec:NAS}, our implemented CNN consists of three layers, with a network width of 32, ReLU activation, batch normalization, and max pooling.

\subsection{Overall Performance}
Table~\ref{t:main results} presents the overall dataset condensation performance. We first describe our evaluation pipeline, followed by a detailed analysis of the results.

\noindent{\textbf{Evaluation Pipeline:}} We evaluate condensation by first training a model on the full dataset and reporting its test accuracy (``Full'') as an upper bound. We then generate condensed datasets controlled by {\em Ratio} (condensed/original size) and {\em SPC} (samples per class). For each setting, we train randomly initialized models on the condensed data and evaluate on the same test set, higher accuracy indicates better knowledge preservation.

\noindent{\textbf{Results Analysis:}} \SC{} consistently outperforms all baselines, with a +17.56\% average gain. It improves over the strongest competitor CondTSC by 6.26\%, and exceeds MTT, SRe$^2$L, and DSA by 7.58\%, 11.12\%, and 14.73\%, respectively (and +19.30\% over Random). With SPC=10, \SC{} retains 88.93\% of full-data performance on average, beating CondTSC (78.45\%), SRe$^2$L (75.11\%), and Random (61.67\%). Even at SPC=1, it preserves 61.84\% of full performance, outperforming CondTSC, SRe$^2$L, and Random by 4.89\%, 15.12\%, and 21.89\%. While both \SC{} and SRe$^2$L use model inversion, \SC{} incorporates shapelets to recover localized discriminative patterns, whereas prior methods mainly capture global temporal characteristics.

To demonstrate that \SC~effectively condenses shapelet knowledge into the synthesized dataset, we conduct a simple yet insightful experiment. Specifically, we train a classifier on shapelet-transformed data from the full dataset and use it to classify the data synthesized by \SC~and other baseline methods. If the synthesized data preserves shapelet patterns, it will be correctly classified by this shapelet-based classifier, and vice versa.

\vspace{-0.05in}
\subsection{Validation of Shapelet Knowledge Preservation}  

\begin{table}[h]
    \centering
    \caption{Validation of shapelet knowledge preservation. Synthesized dataset is evaluated by accuracy via a shapelet-based classifier trained on the full dataset. \SC-$\mathcal{S^*}$ is an ablated variant of \SC, without the shapelet-guided module. Higher accuracy of \SC~indicates its superior ability to preserve shapelet-related knowledge into the condensed dataset.}
    \label{t:shapelet_val}
    \resizebox{0.48\textwidth}{!}{
    \begin{tabular}{lccc}
        \toprule
Condensed Dataset & \SC-$\mathcal{S^*}$ & \SC~& Real Data \\
\midrule
FacesUCR & 52.14 ± 3.71 & \textbf{57.62} ± 0.83 & 67.04 ± 1.93 \\
TP & 69.17 ± 1.44 & \textbf{77.50} ± 2.50 & 76.53 ± 0.00 \\
HAR & 76.11 ± 0.96 & \textbf{92.17} ± 0.99 & 93.09 ± 0.21 \\
Electric & 49.05 ± 5.41 & \textbf{52.38} ± 5.77 & 55.57 ± 0.29 \\
Sleep & 45.87 ± 3.70 & \textbf{48.00} ± 0.00 & 43.25 ± 0.90 \\
Tiselac & 34.07 ± 1.70 & \textbf{63.70} ± 2.57 & 73.42 ± 0.30 \\
Pedestrian & 1.85 ± 0.09 & \textbf{2.24} ± 0.10 & 3.82 ± 0.60 \\
\bottomrule
\end{tabular}
}
\vspace{-0.1in}
\end{table}

As shown in Table~\ref{t:shapelet_val}, the dataset distilled by \SC~consistently achieves higher accuracy than its variant without the shapelet-guided module (\SC-$\mathcal{S^*}$) when evaluated using a shapelet-based classifier. This result highlights \SC’s effectiveness in preserving and condensing shapelet-relevant knowledge into the synthesized dataset.

\begin{table*}[t]
\centering
\caption{Efficiency comparisons conducted during synthesis on the Electric dataset with SPC = 1. 
The results show that \SC~achieves the highest accuracy while maintaining a relatively low computational cost.}
\label{t:efficiency}
\resizebox{0.9\textwidth}{!}{
\begin{tabular}{lccccccccc}
\toprule
 & \multicolumn{4}{c}{\textbf{Image Domain}} 
 & \multicolumn{5}{c}{\textbf{Time-Series Domain}} \\
\cmidrule(lr){2-5} \cmidrule(lr){6-10}
 & DC & DSA & MTT & SRe$^2$L & Random & Herding & K-Center & CondTSC & \SC~ \\
\midrule
Memory Usage (MB) 
& 41.9 & 42.08 & 57.51 & 26.53 
& -- & 31.58 & 31.85 & 102.54 & 31.56 \\

Total Time Cost (s) 
& 198.14 & 595.84 & 1711.72 & 24.71 
& 0.05 & 1.18 & 1.13 & 837.24 & 28.71 \\

Iterations 
& 1000 & 1000 & 2000 & 2000 
& 1 & 25 & 25 & 2000 & 2000 \\

Time per 100 Iterations (s) 
& 19.81 & 59.58 & 85.59 & 1.24 
& 0.05 & 4.73 & 4.52 & 41.86 & 1.44 \\

Accuracy (\%) 
& 45.12 & 46.06 & \underline{52.06} & 49.95 
& 39.29 & 39.55 & 42.86 & 45.32 & \textbf{54.72} \\
\bottomrule
\end{tabular}
}
\end{table*}

\subsection{Efficiency Analysis}

\begin{figure*}[h]
\centering
\includegraphics[width=0.98\linewidth]{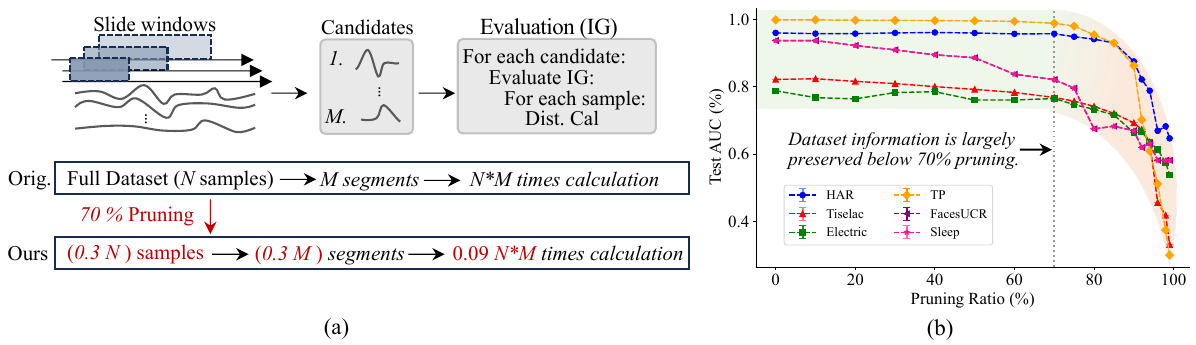}
\vspace{-0.05in}
\caption{High-ratio pruning accelerates shapelet discovery while preserving dataset information. (a) Comparison of shapelet discovery processes with and without high-ratio pruning, showing significant acceleration.
(b) Dataset information remains intact. This is evaluated via classification accuracy of models trained on the pruned dataset. Even with aggressive pruning (up to 70\%), accuracy remains comparable to the original dataset, indicating that essential information is preserved.}
\label{fig: pruning}
\vspace{-0.1in}
\end{figure*}

We evaluate the efficiency of \SC~in comparison to other baselines, focusing on memory usage and time cost during dataset synthesis. The results are in Table~\ref{t:efficiency}. Notably, we exclude the time cost and memory usage for building a teacher model on full dataset (e.g. MTT, SRe$^2$L, CondTSC and \SC), only focusing on data optimization process. Additionally, since convergence iterations vary across methods, we report both total time cost and time per 100 iterations to provide a more comprehensive evaluation.

The results show that \SC~achieves highest performance while spending the second-lowest memory usage of 31.56 MB and the second-lowest time 28.71 s for sample synthesis. 
Compared to CondTSC, which is also designed for time series data, our approach reduces 69\% memory usage and 96.5\% total time cost. This efficiency derives from its decoupled optimization process.
Compared to SRe$^2$L, our approach achieves superior performance by 9.55\% through slightly compromising in memory usage and time cost to integrate informative shapelet-related calculation in our framework.
Among all methods, Core-set Selection based methods require the least time but deliver the poorest performance, making them suitable for scenarios where fast generation is prioritized over data quality.

\subsection{Ablations}

\noindent{\bf Impact of Shapelet Size.}
We study the effect of the shapelet set size $|\mathcal{S}^*|$ on condensation performance in \SC. 
Table~\ref{t:num_sl} reports the classification accuracy under different shapelet sizes. 
Performance varies across datasets and condensation ratios, and no single shapelet size consistently yields the best results.
This is expected, since the appropriate number of shapelets depends on both the dataset and the compression ratio.

In practice, moderate to large shapelet sizes (e.g., $|\mathcal{S}^*| \in \{10, 20\}$) provide competitive performance across datasets.

\begin{table}[h]
    \centering
    \caption{
    Impact of shapelets size to condensation accuracy (\%). Best Accuracy shown in bold, the second shown in underline. }
    \label{t:num_sl}
    \resizebox{0.48\textwidth}{!}{
    \begin{tabular}{ccccccc}
    \toprule
    Dataset & Ratio (\%) & SPC & $|\mathcal{S}^*|=5$ & $|\mathcal{S}^*|=10$ & $|\mathcal{S}^*|=15$ & $|\mathcal{S}^*|=20$ \\
    \midrule
\multirow{3}{*}{HAR} & 0.01 & 1 & 66.22 & 76.44 & 61.68 & 71.52 \\
 & 0.05 & 5 & 85.24 & 89.10 & 75.14 & 87.95 \\
 & 0.11 & 10 & 88.90 & 92.08 & 82.06 & 89.68 \\
\midrule
\multirow{2}{*}{TP} & 0.07 & 1 & 36.46 & 28.62 & 44.19 & 45.95 \\
 & 0.71 & 10 & 78.37 & 76.77 & 83.76 & 90.94 \\
\midrule
\multirow{2}{*}{Sleep} & 0.20 & 10 & 43.90 & 41.19 & 35.28 & 47.08 \\
 & 0.98 & 50 & 68.21 & 66.22 & 66.17 & 67.53 \\
\bottomrule
\end{tabular}
}
\vspace{-0.15in}
\end{table}

\begin{table*}[h]
\centering
\caption{Neural Architecture Search for 432 convolutional networks on the HAR dataset.
\SC~achieves performance closest to that trained on the full dataset.}
\label{t:nas}
\resizebox{0.88\textwidth}{!}{
\begin{tabular}{lccccccccc c}
\toprule
 & \multicolumn{4}{c}{\textbf{Image Domain}} 
 & \multicolumn{5}{c}{\textbf{Time-Series Domain}} 
 & \textbf{Full} \\
\cmidrule(lr){2-5} \cmidrule(lr){6-10}
 & DC & DSA & MTT & SRe$^2$L 
 & Random & Herding & K-Center & CondTSC & \SC~ 
 &  \\
\midrule
Accuracy (\%) 
& 84.02 & 87.28 & 93.76 & 94.88 
& 81.45 & 83.34 & 82.88 & 93.39 & \textbf{95.26} 
& 96.12 \\

Training samples 
& 60 & 60 & 60 & 60 
& 60 & 60 & 60 & 60 & 60 
& 5878 \\
\bottomrule
\end{tabular}
}
\vspace{-0.08in}
\end{table*}

\noindent{\bf Impact of Pruning Ratio.}
\label{sec:PR}
We study high-ratio pruning as an acceleration strategy for shapelet discovery, reducing computation while preserving discriminative information. Unlike Table~\ref{t:main results}, we do not apply class-balanced resampling.

Fig.~\ref{fig: pruning}(b) shows that pruning up to 70\% of the training data yields classification accuracy close to using the full dataset, suggesting that most discriminative signals remain under moderate pruning. Consistently, Fig.~\ref{fig: pruning}(a) indicates that such pruning substantially speeds up shapelet discovery by reducing candidate segments and similarity evaluations. In particular, the evaluation cost drops from $(N\!\times\!M)$ to $(0.09N\!\times\!M)$.

However, extreme pruning (e.g., below 10\%) causes a sharp accuracy drop, revealing the limits of aggressive compression and motivating condensation methods that preserve critical patterns in ultra-compact subsets.

\section{Downstream: Neural Architecture Search}
\label{sec:NAS}

Our \SC~produces a highly information-dense condensed dataset that is far smaller than the original, substantially reducing downstream compute. We illustrate this on Neural Architecture Search (NAS), which is costly because it repeatedly trains many candidate networks. Evaluating $k$ architectures costs $k,f(n_{\text{full}})$ on the full dataset, but drops to $k,f(n_{\text{condensed}})$ with $n_{\text{condensed}} \ll n_{\text{full}}$, enabling efficient search while preserving evaluation fidelity.

Following \citet{condtsc}, we generate condensed data with nine methods and run NAS over 432 CNN architectures on HAR (search space details in Appendix Table~\ref{t:nas_settings}). As shown in Table~\ref{t:nas}, \SC~reaches 95.26\% top accuracy, 99.11\% of full-data performance (96.12\%), demonstrating a data-efficient proxy for large-scale NAS with minimal accuracy loss.

\section{Related Work}
\subsection{Dataset Condensation}
Dataset condensation or distillation synthesizes a compact training set that preserves the key information of the original data, reducing storage and training cost. Existing methods mainly fall into \emph{bi-level} \cite{wang2018dataset, dcdm, dsa, mtt} and \emph{single-level} \cite{sre2l,yindataset,shen2025delt,shao2024generalized,cui2025fadrm} optimization.

\noindent\textbf{Bi-level methods.}
Bi-level approaches formulate condensation as a nested problem, optimizing synthetic data to improve a meta-objective measured on real data. Gradient-matching methods such as DC \cite{dcdm} align gradients between synthetic and real batches, and DSA \cite{dsa} further improves performance with differentiable augmentation. Trajectory-matching methods match training dynamics, e.g., MTT \cite{mtt} matches parameter trajectories and TESLA \cite{tesla} extends to longer horizons. While effective, bi-level methods are typically computationally heavy due to many inner-loop updates.

\noindent\textbf{Single-level methods.}
Single-level methods avoid nested optimization by directly matching distributions. SRe$^2$L \cite{sre2l} improves efficiency via model inversion, aligning BatchNorm running statistics. Following this line, we adopt a single-level model-inversion strategy for scalable condensation. DM \cite{zhao2023dataset} matches feature statistics (e.g., moments) but can still be expensive when repeated across random networks.

\noindent\textbf{Core-set selection.}
Core-set methods~\cite{tsang2005core, har2005smaller, farahani2009facility,chen2012super, sener2017active, rebuffi2017icarl, castro2018end} select representative real samples without synthesis. Examples include K-Center~\cite{har2005smaller, farahani2009facility, malkomes2015fast, sener2017active} for coverage and Herding~\cite{welling2009herding} for matching dataset moments. Although not condensation in the strict sense, they provide strong compact baselines.

\vspace{-0.05in}
\subsection{Time Series Dataset Condensation}
\vspace{-0.05in}

Despite broad progress on images and graphs, time series condensation remains underexplored. CondTSC \cite{condtsc} observes that image-centric methods transfer poorly to time series, and adapts the bi-level MTT \cite{mtt} framework with frequency-domain cues. However, it mainly emphasizes global structure, often missing \emph{localized} discriminative patterns, and it inherits the high cost of bi-level optimization. This motivates more efficient time-series-specific condensation.

\noindent\textbf{Shapelets as intrinsic knowledge.}
 Shapelets are class-discriminative subsequences that capture essential local temporal patterns. Since their introduction for time series classification \cite{ye2009time}, shapelets have been widely used and extended \cite{grabocka2014learning, hou2016efficient, li2021shapenet, le2024shapeformer}. Yet, they have not been systematically leveraged for dataset condensation.
\vspace{-0.03in}
To bridge this gap, we propose \SC{}, a shapelet-guided \emph{single-level} condensation framework tailored to time series. By explicitly preserving local shapelet patterns alongside global temporal structure, \SC{} achieves state-of-the-art performance with improved efficiency.

\vspace{-0.05in}
\section{Conclusion}
We proposed \SC, a novel shapelet-guided dataset condensation framework that effectively incorporates local discriminative patterns into the condensation process for time series data. By integrating shapelet-guided optimization with global structure modeling, our method preserves both local and global temporal dynamics. Extensive experiments on seven public datasets demonstrate consistent and significant performance improvements over existing methods, achieving state-of-the-art results. This work establishes a promising paradigm for efficient temporal data modeling in resource-constrained settings.

\section*{Impact Statement}
This work introduces a method for efficiently condensing large-scale time series datasets, advancing scalable and efficient ML research while reducing computational cost, memory usage, and energy consumption, contributing to more sustainable machine learning. Overall, the societal impact of this work is primarily positive, however, condensed datasets may introduce distortions or bias, so they should be used with caution.

\bibliography{example_paper}
\bibliographystyle{icml2026}

\newpage
\appendix
\onecolumn
\section*{\Large Appendix}

\section{Datasets}
We evaluate our method on seven public time series datasets, including FacesUCR\footnote{\url{https://www.timeseriesclassification.com/index.php}\label{data1}}, TwoPatterns (TP)\textsuperscript{\ref{data1}}, Activity Recognition (HAR)\cite{anguita2013public}, ElectricDevices (Electric)\textsuperscript{\ref{data1}}, Sleep\textsuperscript{\ref{data1}}, Tiselac\textsuperscript{\ref{data1}}, and Pedestrian\footnote{\url{https://huggingface.co/monster-monash}\label{data2}}. Given that dataset condensation is particularly valuable for large-scale scenarios, we include Pedestrian for its substantial size. 
These datasets are from diverse fields, including facial recognition, synthetic pattern analysis, human activity monitoring, electric device measurement, medical diagnosis, sleep stage classification, and environmental sensing, providing a comprehensive benchmark for assessing the effectiveness of our approach.
Detailed statistics and split ratio of these datasets are summarized in Table~\ref{t:dataset}. 

Since dataset condensation is particularly valuable for large-scale datasets, we employ an increased split ratio (compared to the original split ratio in UEA datasets) to expand the training set size. 
During the partitioning, we ensure that the class distributions remain consistent across the training, validation, and testing sets. Additionally, we observe that certain datasets, such as FacesUCR, Sleep, Tiselac, and Pedestrian, exhibit class imbalance. In line with established practices in time series classification \cite{mohammadi2024deep, middlehurst2024bake}, we apply oversampling \cite{chawla2002smote} to ensure class balance in preprocessing. 

\begin{table}[h]
    \centering
    \caption{The statistics and split ratio of datasets.}
    \label{t:dataset}
    \resizebox{0.65\textwidth}{!}{ 
    \begin{tabular}{lccccccc}
    \toprule
        Dataset & FacesUCR & TP & HAR & Electric & Sleep & Tiselac & Pedestrian \\ \midrule
        Timesteps & 131 & 128 & 128 & 96 & 3000 & 23 & 24 \\ \midrule
        Variables & 1 & 1 & 9 & 1 & 1 & 10 & 1 \\ \midrule
        Classes & 14 & 4 & 6 & 7 & 5 & 9 & 82 \\ \midrule
        Imbalance Ratio & 0.14 & 0.92 & 0.72 & 0.29 & 0.16 & 0.09 & 0.02 \\ \midrule
        \# Trainset (orig.) & 1344 & 1249 & 5878 & 7807 & 25609 & 79747 & 151663 \\ \midrule
        \# Trainset (bal.) & 2744 & 1304 & 6660 & 14049 & 53870 & 144000 & 320866 \\ \midrule
        \# Valset & 457 & 1876 & 1477 & 1121 & 7791 & 9971 & 18998 \\ \midrule
        \# Testset & 449 & 1875 & 2944 & 7709 & 8908 & 9969 & 18960 \\ 
    \bottomrule
    \end{tabular}
    }
\end{table}

\section{Baselines}

To evaluate \SC~on time series dataset condensation, we compare it against 10 baselines spanning both image and time series domains. 
These baselines are representative state-of-the-art methods for dataset condensation.

\begin{itemize}
    \item \textbf{Image-domain methods:} Originally developed for images, they include DC \cite{dcdm} and MTT \cite{mtt}, which use one-step and trajectory gradient matching respectively; DSA \cite{dsa}, which applies siamese augmentation to synthesize informative samples; and SRe$^2$L \cite{sre2l}, which performs dataset distillation via model inversion constrained by batch normalization statistics.
    Among these, SRe$^2$L is the most relevant competitor, as it shares the model inversion paradigm with \SC.
    
    \item \textbf{Time series-domain methods:} These include Random sampling, K-Means, and Herding, which are general-purpose core-set selection approaches applicable to time series, as well as CondTSC \cite{condtsc}, a recently proposed dataset condensation method designed for time series.
\end{itemize}

Compared to these baselines, the key design of \SC~is its integration of domain-specific information for time series.

\section{Metrics}

The effectiveness of dataset condensation methods is assessed based on the performance of models trained on the synthesized datasets. For condensing time series classification datasets, we use Accuracy to evaluate the model performance:
\[
\text{Accuracy} = \frac{\sum_{i=1}^C \text{Correct}_i}{\sum_{i=1}^C \text{Total}_i},
\]
where $C$ denotes the total number of classes, $\text{Correct}_i$ is the number of correctly predicted samples for class $i$, and $\text{Total}_i$ is the total number of samples for class $i$.

To further evaluate the dataset condensation methods, we compare the performance of a model trained on the synthesized dataset $\mathcal{C}$ to that of a model trained on the full dataset $\mathcal{T}$. Let $\theta_{\mathcal{C}}$ and $\theta_{\mathcal{T}}$ denote the parameters of the model $\mathcal{F}$ trained on $\mathcal{C}$ and $\mathcal{T}$, respectively. The ratio of their accuracies serves as the evaluation metric for the distillation process:
\[
\text{Accuracy Ratio} = \frac{\text{Accuracy}(\theta_{\mathcal{C}})}{\text{Accuracy}(\theta_{\mathcal{T}})}.
\]

\section{More Details of NAS}
We define a discrete NAS search space for CNN architectures, covering five architectural dimensions: network depth, width, normalization, activation, and pooling.
The depth controls the number of convolutional blocks, while the width determines the number of channels in each block.
Different normalization and activation functions are included to account for their impact on training stability and non-linearity.
Pooling operations are optionally applied to model temporal downsampling in time series.
The candidate values for each dimension are listed in Table~\ref{t:nas_settings}, resulting in a total of 432 candidate architectures.

\begin{table}[tp]
    \centering
    \caption{Searching space of NAS for CNN networks. In total, there are 432 candidate architectures.}
    \label{t:nas_settings}
    \resizebox{0.5\textwidth}{!}{
    \begin{tabular}{lc}
        \toprule
        Searching Dimension & Candidate Values \\
                \midrule
        Depth & 2, 3, 4 \\
                \midrule
        Width & 32, 64, 128, 256\\
                \midrule
        Normalization & None, BatchNorm, InstanceNorm, LayerNorm \\
                \midrule
        Activation & Sigmoid, ReLU, LeakyReLU \\
                \midrule
        Pooling & None, Max, Mean\\
        \bottomrule
    \end{tabular}
    }
\end{table}

\section{ShapeCond with Alternative Backbone Architectures}
In the main experiments, ShapeCond is instantiated with a CNNBN backbone to align with prior dataset condensation studies (e.g., CondTSC) and ensure fair comparison. 
Importantly, the proposed framework is backbone-agnostic and not inherently tied to this specific architecture.

To examine its flexibility, we further instantiate ShapeCond with a Transformer-based backbone and evaluate it across multiple datasets and condensation ratios. 
As shown in Table~\ref{t:transformer_results}, ShapeCond maintains strong performance under this alternative architecture and achieves competitive, and in many cases superior, results compared to existing methods. 
These results suggest that the effectiveness of ShapeCond does not depend on a particular backbone choice, but rather arises from its optimization design.

\begin{table*}[tp]
\centering
\caption{Overall performance (Accuracy \%) of dataset condensation methods when ShapeCond is instantiated with a Transformer backbone.  
All methods are evaluated using the same training and evaluation protocol as in the CNNBN experiments. 
All reported values are averaged over three runs, and the best accuracy is highlighted in bold.}
\label{t:transformer_results}
\resizebox{.91\textwidth}{!}{
\begin{tabular}{lcccccccccccc}
\toprule
Dataset & Ratio(\%) & SPC & Random & Herding & K-Center & DC & DSA & MTT & SRe$^2$L & CondTSC & \SC{} (Ours) & Full \\
\midrule
\multirow{3}{*}{HAR} 
 & 0.1 & 1  
 & 58.02 & 59.99 & 63.99 & 67.97 & 68.53 & 65.86 & 68.52 & 66.20 & \textbf{76.44} & \multirow{3}{*}{94.24} \\
 & 0.5 & 5  
 & 61.65 & 60.46 & 61.24 & 74.78 & 74.42 & 72.89 & 73.83 & 79.37 & \textbf{86.92} &  \\
 & 1.0 & 10 
 & 62.06 & 62.68 & 69.97 & 76.03 & 79.70 & 81.60 & 83.44 & 83.45 & \textbf{89.22} &  \\
\midrule
\multirow{3}{*}{Electric} 
 & 0.1 & 1  
 & 35.54 & 29.33 & 29.91 & 27.49 & 30.23 & 30.69 & 32.86 & 28.86 & \textbf{35.36} & \multirow{3}{*}{53.25} \\
 & 0.4 & 5  
 & 29.56 & 24.56 & 29.01 & 30.25 & 32.54 & 35.12 & 34.77 & 36.98 & \textbf{40.59} &  \\
 & 0.9 & 10 
 & 40.89 & 35.93 & 39.24 & 32.75 & 33.96 & 37.06 & 37.45 & 38.37 & \textbf{42.55} &  \\
\bottomrule
\end{tabular}
}
\end{table*}

\section{More Initialization Methods}
We study two initialization schemes for dataset condensation: random and instance initialization.
Random initialization draws synthetic samples from a normal distribution, while instance initialization initializes the synthetic set by randomly selecting samples from the original dataset.
Table~\ref{t:init} presents classification accuracy (\%) for random and instance initialization on HAR, Electric, and TP.

\begin{table}[tp]
    \centering
    \caption{Effect of random and instance initialization on dataset condensation performance.}
    \label{t:init}
\resizebox{0.4\textwidth}{!}{
\begin{tabular}{ccccc}
\toprule
 & Ratio(\%) & SPC & Random Init & Instance Init \\
\midrule
\multirow{3}{*}{HAR} & 0.1 & 1 & 76.44 & 65.00 \\
 & 0.5 & 5 & 89.10 & 84.50 \\
 & 1.0 & 10 & 92.08 & 89.30 \\
\midrule
\multirow{3}{*}{Electric} & 0.1 & 1 & 54.72 & 53.08 \\
 & 0.4 & 5 & 65.38 & 60.04 \\
 & 0.9 & 10 & 68.30 & 59.68 \\
\midrule
\multirow{2}{*}{TP} & 0.3 & 1 & 45.95 & 42.32 \\
 & 1.6 & 5 & 60.29 & 51.28 \\
 & 3.2 & 10 & 90.94 & 69.56 \\
 \bottomrule
    \end{tabular}
    }
\end{table}

\section{Theoretical Efficiency Analysis of Shapelet Discovery}
\label{app:complexity_shapelet}

\noindent\textbf{Notation.} We use \(N\) for the number of time series, \(L\) for the sequence length, \(M\) for the total number of candidate subsequences generated by the conventional pipeline (before any pruning), \(p\) for the pruning ratio on samples (so that \((1-p)N\) samples remain), and \(K\) for the number of shapelets used for representation. We assume the candidate set size \(M\) scales as \(M= \Theta(NL^2)\) in the conventional sliding-window enumeration; \(|s|\) denotes a candidate length (assumed bounded by a constant).

\subsection{Conventional pipeline: precise counting and complexity}
The conventional pipeline enumerates all subsequences as candidates. Denote by \(M\) the total number of such candidates:
\begin{equation}
M \;=\; \sum_{i=1}^{N}\sum_{l=L_{\min}}^{L_{\max}} (L-l+1) \;=\; \Theta(NL^2),
\end{equation}

For a candidate \(s\) (of length \(|s|\) assumed \(O(1)\)), its distance to a series \({\bf x}\) is
\begin{equation}
D(s,{\bf x}) \;=\; \min_{t\in[1,\,L-|s|+1]} \| s - {\bf x}_{t:t+|s|-1}\|_2,
\end{equation}
which requires scanning \(\Theta(L)\) alignment positions. Thus the cost of a single candidate-vs-series distance evaluation is \(O(L)\).

During discovery, each candidate is compared to all \(N\) series, so the number of (alignment) operations is
\[
\text{Conventional distance ops} \;=\; M \cdot N \cdot \Theta(L).
\]
Using \(M=\Theta(NL^2)\) this yields the commonly cited overall complexity
\[
\boxed{\;\text{Conventional discovery: } \; O(M \cdot N \cdot L) \;=\; O(N^2 L^3)\; }.
\]

\subsection{Our accelerated pipeline: counting and complexity}
We apply two complementary accelerations:

\textbf{(A) Sample pruning.} We perform discovery on a random subset of samples, keeping fraction \((1-p)\) of the dataset. Under the mild assumption that candidate generation scales with the number of samples, the number of candidates retained becomes
\[
M_{\text{post}} \;=\; (1-p)\,M.
\]
(The same \(p\) denotes the fraction of samples removed; this yields proportional reduction in generated candidates in the pruning stage.)

\textbf{(B) Position-constrained distance search.} For a candidate extracted at a nominal position \(t\), rather than scanning the entire sequence we only search inside a local window \([t-W,\,t+W]\) of constant size \(2W+1\). If \(W\) and \(|s|\) are constants independent of \(L\), the per-candidate-per-series distance cost becomes constant:
\[
\text{per comparison cost} = O(W\cdot |s|)=O(1),
\]

Combining (A) and (B), the total number of alignment operations in our pipeline is
\[
\text{Proposed distance ops} \;=\; M_{\text{post}} \cdot N_{\text{used}} \cdot O(1),
\]
where \(N_{\text{used}}=(1-p)N\) is the number of samples used during discovery. Thus
\[
\text{Proposed distance ops} \;=\; (1-p)M \cdot (1-p)N \cdot O(1) \;=\; (1-p)^2 M N \cdot O(1),
\]
Writing this in the same high-level form as the conventional expression gives
\[
\boxed{\;\text{Proposed discovery: } \; O\big((1-p)^2 M N\big) \;=\; O\big((1-p)^2 N^2 L^2\big)\;}.
\]
where we used \(M=\Theta(NL^2)\) to expand in \(N,L\).

\subsection{Representation (time series $\to$ shapelet features)}
After discovery we typically compute shapelet-based features for each series using the discovered \(K\) shapelets. In the conventional approach each distance \(D(s,{\bf x})\) requires a full-sequence scan, so
\[
\text{Conventional representation cost} \;=\; N \cdot K \cdot \Theta(L) \;=\; O(N K L).
\]
If we apply the same position-constrained evaluation at representation time (i.e., compute distances only in a small window), the per-distance cost becomes \(O(1)\) and representation cost reduces to
\[
\text{Proposed representation cost} \;=\; O(N K).
\]

\subsection{Complexity ratio and practical savings}
\label{app_cost}
Compare total elementary alignment operations (conventional vs proposed discovery):
\[
\frac{\text{Conventional ops}}{\text{Proposed ops}} \;=\; \frac{M N \cdot L}{(1-p)^2 M N \cdot c} \;=\; \frac{L}{c(1-p)^2},
\]
where \(c=2W+1\) is the number of alignment positions checked per comparison in the position-constrained search (a small constant). Thus the asymptotic \emph{reduction factor} is proportional to \(L/(1-p)^2\): for long sequences (\(L\) large) and aggressive pruning (large \(p\)), the savings are very large.

\paragraph{Concrete example.} Suppose \(N=10{,}000\), \(L=3000\), pruning ratio \(p=0.7\) (so \(1-p=0.3\)), and we check \(c=3\) positions per local search (position and ±1). Then the ratio of elementary alignment operations (conventional : proposed) is
\[
\frac{L}{c(1-p)^2} \;=\; \frac{3000}{3 \times 0.3^2} \;=\; \frac{3000}{3 \times 0.09} \;=\; \frac{3000}{0.27} \approx 11{,}111.
\]
That is, under these parameters the proposed method performs on the order of \(1.1\times 10^{4}\) fewer alignment operations than the conventional full-scan discovery (the absolute counts are proportional to \(M N\) and cancel out in the ratio).

\subsection{Concluding remarks}
\begin{itemize}
  \item The conventional end-to-end discovery scales as \(O(M N L)\) (with \(M=\Theta(NL^2)\) giving \(O(N^2 L^3)\)), which is prohibitive for large \(N\) or long \(L\).
  \item By combining (i) sample/candidate pruning and (ii) position-constrained local search, \SC{} reduces the discovery cost to \(O((1-p)^2 M N)\) (which expands to \(O((1-p)^2 N^2 L^2)\) under \(M=\Theta(NL^2)\)), and can reduce representation cost from \(O(N K L)\) to \(O(N K)\) when local search is used at representation time.
  \item The practical savings scale roughly as \(L/(c(1-p)^2)\); for realistic \(L\) and moderate-to-high pruning \(p\) this yields orders-of-magnitude reductions in alignment work.
\end{itemize}

\end{document}